\documentclass[runningheads]{llncs}

\usepackage{eccv}
\usepackage{geometry}
\usepackage{eccvabbrv}
\usepackage{multirow,multicol, makecell, booktabs}
\usepackage{pifont}
\usepackage{graphicx}
\usepackage{booktabs}

\usepackage[accsupp]{axessibility}  

\usepackage{hyperref}

\usepackage{orcidlink}

\begin{document}

\title{Decoding Matters: Efficient Mamba-Based Decoder with Distribution-Aware Deep Supervision for Medical Image Segmentation} 

\titlerunning{Decoding Matters}

\author{Fares Bougourzi\inst{1}\orcidlink{0000-0001-5077-4862} \and
Fadi Dornaika\inst{2}\orcidlink{0009-0003-0697-9301} \and
Abdenour Hadid\inst{3}\orcidlink{0000-0001-9092-735X}}

\authorrunning{F.~Bougourzi et al.}

\institute{
IEMN UMR CNRS 8520, Université Polytechnique Hauts-de-France, 59313 Valenciennes, France,
\email{faresbougourzi@gmail.com}\\
\and
University of the Basque Country (UPV/EHU), Donostia-San Sebastián, Spain, \email{fdornaika@gmail.com}\\
\and
Sorbonne Center for Artificial Intelligence, Sorbonne University Abu Dhabi, Abu Dhabi, UAE,
\email{abdenour.hadid@ieee.org} 
}

\maketitle

\begin{abstract}
Deep learning has achieved remarkable success in medical image segmentation, often reaching expert-level accuracy in delineating tumors and tissues. However, most existing approaches remain task-specific, showing strong performance on individual datasets but limited generalization across diverse imaging modalities. Moreover, many methods focus primarily on the encoder, relying on large pretrained backbones that increase computational complexity. In this paper, we propose a decoder-centric approach for generalized 2D medical image segmentation. The proposed \textbf{Deco-Mamba} follows a U-Net-like structure with a Transformer-CNN-Mamba design. The encoder combines a CNN block and Transformer backbone for efficient feature extraction, while the decoder integrates our novel Co-Attention Gate (CAG), Vision State Space Module (VSSM), and deformable convolutional refinement block to enhance multi-scale contextual representation. Additionally, a windowed distribution-aware KL-divergence loss is introduced for deep supervision across multiple decoding stages. Extensive experiments on diverse medical image segmentation benchmarks yield state-of-the-art performance and strong generalization capability while maintaining moderate model complexity. The source code will be released upon acceptance.
  \keywords{Segmentation \and Mamba \and Transformer \and Medical Imaging}
\end{abstract}

\section{Introduction}
\label{sec:intro}

Medical image segmentation (MIS) plays a pivotal role in computer-assisted diagnosis and treatment planning, aiming to assign semantic labels to each pixel of a medical image. Recent advances in deep learning have significantly improved segmentation accuracy, driven by architectural innovations that enhance feature representation and contextual modeling~\cite{yao2024cnn, zhang2025advances, bougourzi2025recent}.

The foundation of modern MIS can be traced back to the introduction of U-Net~\cite{ronneberger2015u}, a CNN-based architecture that marked a major breakthrough. Its encoder--decoder design with skip connections preserves high-resolution spatial details while enabling multi-scale feature extraction. This paradigm became the cornerstone for subsequent architectures, inspiring numerous variants such as Attention U-Net~\cite{oktay2018attention}, U-Net++~\cite{zhou_unet_2018}, M-Net~\cite{fu2018joint}, ResNet-U-Net~\cite{yu2019robust}, and PDAtt-U-Net~\cite{bougourzi2023pdatt}. These models demonstrated strong performance across imaging modalities including MRI, CT, and ultrasound by effectively capturing local patterns through convolutional operations. However, CNN-based approaches inherently struggle to model long-range dependencies due to their limited receptive field, resulting in suboptimal contextual reasoning and reduced generalization across imaging modalities and tasks.

To overcome these limitations, Transformers~\cite{dosovitskiy2020vit, yao2024cnn} have been introduced into vision tasks for their self-attention mechanism, which effectively models global dependencies. In medical image segmentation, architectures such as TransUNet~\cite{chen2021transunet}, Swin-UNet~\cite{liu_swin_2021}, UCTransNet~\cite{wang2022uctransnet}, Cascaded-MERIT~\cite{rahman2024multi}, and PAG-TransYNet~\cite{bougourzi2024rethinking} incorporated Transformer backbones within encoder--decoder frameworks. These models achieved state-of-the-art (SOTA) performance across multiple benchmarks by capturing long-range spatial dependencies and enhancing global feature aggregation. Nevertheless, the quadratic complexity of self-attention with respect to sequence length makes these models computationally expensive and memory-intensive, limiting their scalability to high-resolution medical images and practical deployment.

\begin{figure}[h]
\begin{center}
\includegraphics[width=0.7\linewidth, height=2.2in]{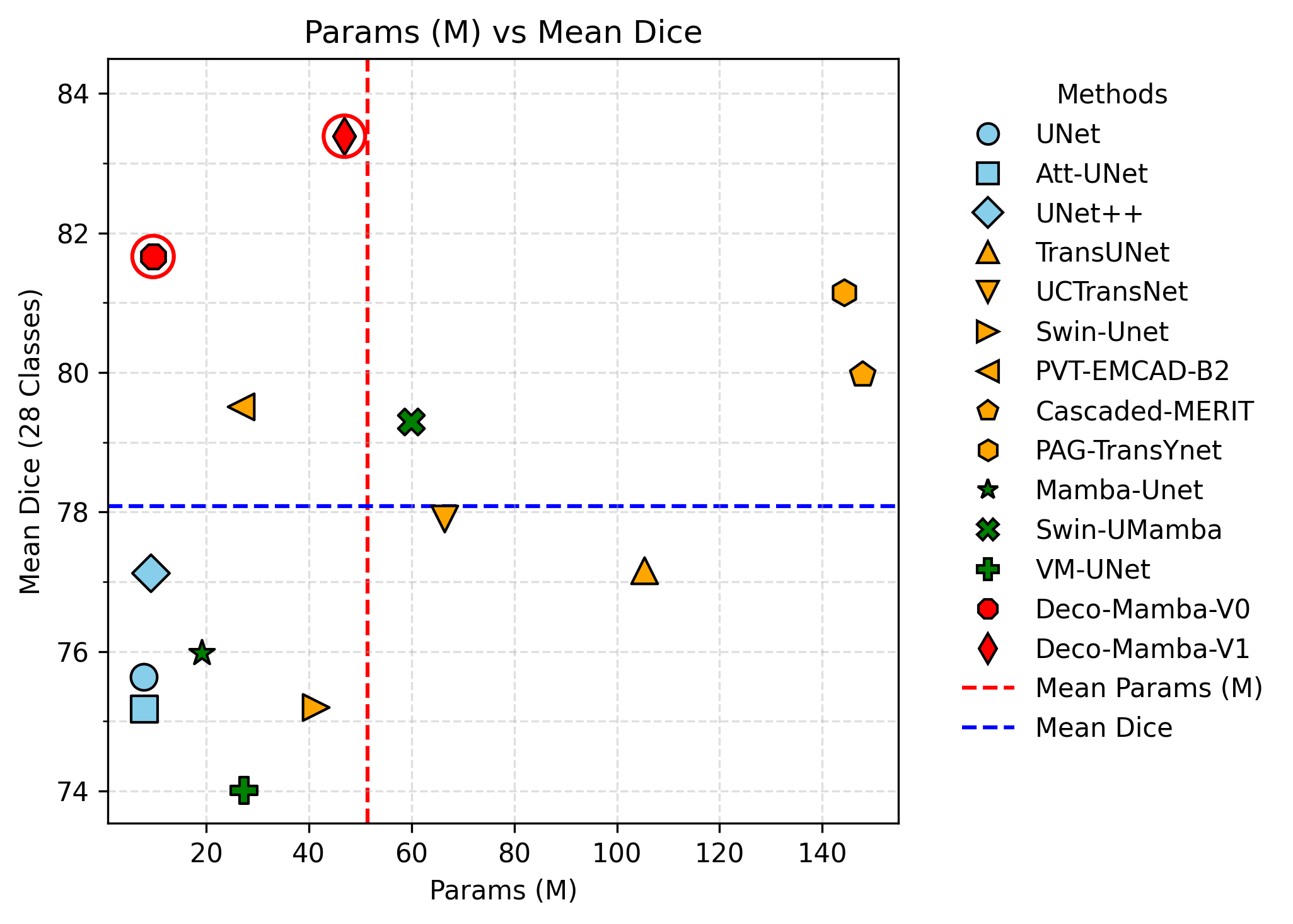} 
\caption{Comparison of Mean Dice Scores of SOTA methods with Respect to Model Complexity in terms of Parameters (M). Our proposed approaches are denoted \textbf{Deco-Mamba-V\textsubscript{0}} and \textbf{Deco-Mamba-V\textsubscript{1}}. }
\label{fig:params}
\end{center}
\end{figure}

\begin{figure}[h]
\begin{center}
\includegraphics[width=0.7\linewidth, height=2.2in]{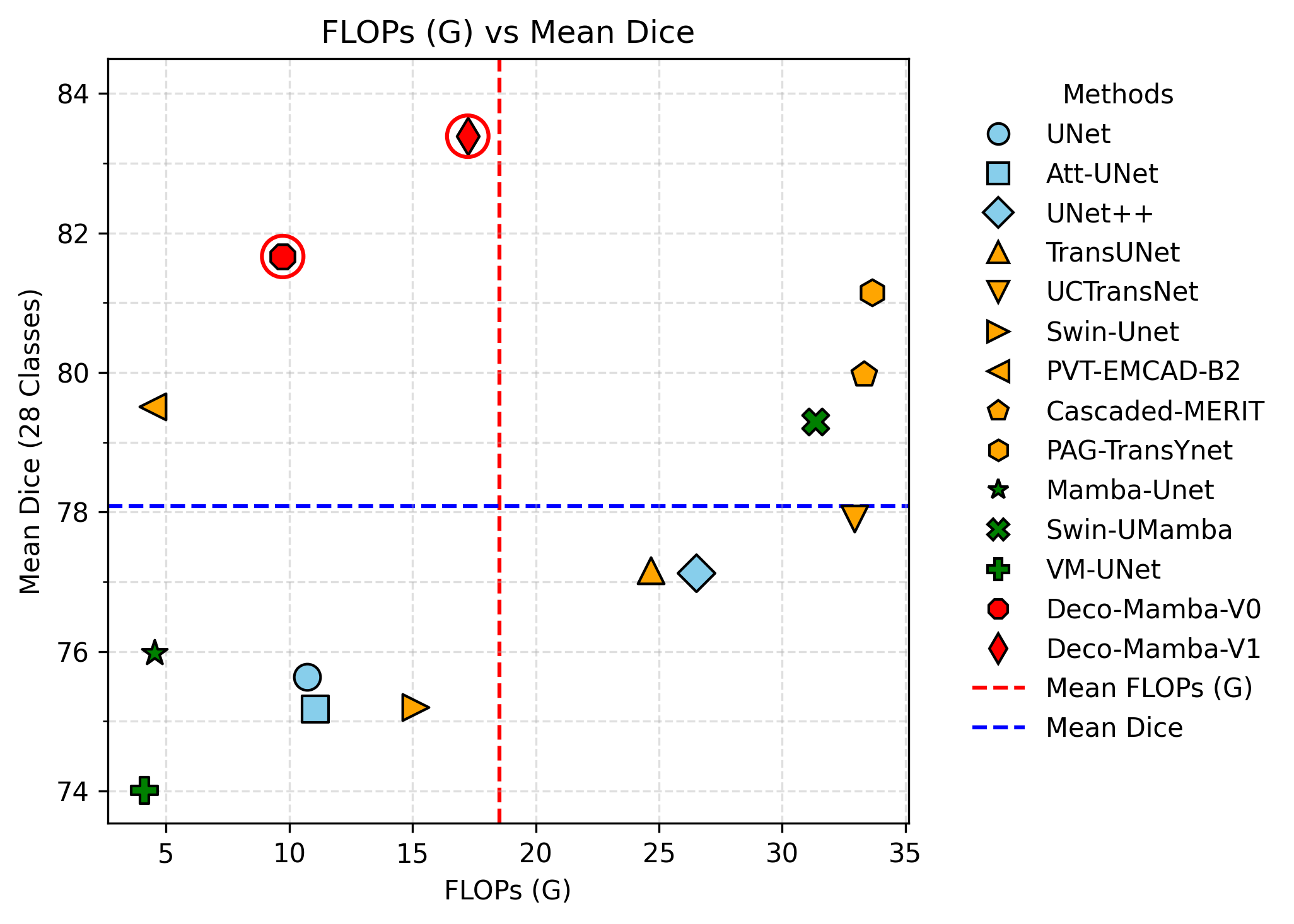} 
\caption{Comparison of Mean Dice Scores with Respect to Model Complexity in terms of FLOPs (G). Our proposed approaches are denoted \textbf{Deco-Mamba-V\textsubscript{0}} and \textbf{Deco-Mamba-V\textsubscript{1}}.}
\label{fig:flops}
\end{center}
\end{figure}

Recently, State Space Models (SSMs)~\cite{gu2021efficiently, gu2024mamba}, and in particular the Mamba architecture~\cite{gu2021efficiently, gu2024mamba}, have emerged as efficient alternatives to Transformers. Mamba replaces self-attention with selective state-space updates that model long-range dependencies in linear time and space complexity~\cite{liu2024vmamba}. This innovation has sparked a new research direction in vision, leading to several hybrid Mamba-based segmentation frameworks such as SliceMamba~\cite{fan2025slicemamba}, Swin-UMamba~\cite{linguraru_swin-umamba_2024}, Mamba-UNet~\cite{wang2024mamba}, and U-Mamba~\cite{U-Mamba}. These models successfully reduce computational overhead while maintaining competitive accuracy. However, most of these methods have been validated on a limited number of imaging modalities and focus primarily on enhancing the encoder through Mamba-based backbones. Consequently, these approaches often struggle to generalize across heterogeneous datasets and diverse imaging conditions, as illustrated in Figures~\ref{fig:params} and~\ref{fig:flops}.

While recent studies have primarily focused on encoder improvements~\cite{linguraru_swin-umamba_2024, ruan2024vm, bougourzi2024rethinking, zhang2023st, chen2021transunet}, the decoder has received comparatively little attention, despite its crucial role in reconstructing fine-grained spatial details~\cite{rahman2024emcad, rahman2024multi}. Although powerful encoders can extract rich semantic representations, under-designed decoders may struggle to accurately recover object boundaries and contextual structures during upsampling. This motivates a shift toward a \emph{decoder-centric} design that emphasizes multi-scale refinement, adaptive attention mechanisms, and distribution-aware supervision to improve reconstruction fidelity and generalization across diverse domains.

Existing enriched decoder architectures often employ cascaded decoder structures aligned with cascaded encoders, which substantially increase computational complexity~\cite{rahman2024multi}. EMCAD~\cite{rahman2024emcad}, in contrast, integrates attention mechanisms with lightweight convolutional blocks and conventional deep supervision but does not explicitly model long-range dependencies. To address these limitations, \textbf{Deco-Mamba} introduces a novel co-attention gating mechanism that effectively fuses multi-level features while capturing long-range dependencies directly within the decoder. Additionally, we propose a distribution-aware deep supervision strategy to improve boundary precision and enhance robustness under domain shifts.

In this work, we propose \textbf{Deco-Mamba}, a \emph{decoder-centric Mamba-based architecture} for generalized 2D medical image segmentation. Our model follows a U-Net-like structure with a hybrid Transformer--CNN encoder for efficient feature extraction. The decoder incorporates three key components: (1) a novel Co-Attention Gate to adaptively refine skip and decoder features, (2) a Vision State Space Module for efficient multi-scale context modeling, and (3) a deformable convolutional refinement layer to recover structural precision. Additionally, we introduce a \textbf{windowed distribution KL-divergence loss} for multi-scale deep supervision, encouraging distributional consistency across decoding stages and mitigating information loss caused by resizing operations.

Our main contributions can be summarized as follows:
\begin{itemize}
    \item We propose a \textbf{decoder-centric Mamba-based architecture} that achieves SOTA performance with \textbf{improved computational efficiency} compared to existing methods across multiple medical imaging modalities.
    \item We introduce a Co-Attention Gate and a Vision State Space Module for adaptive multi-scale feature modeling and enhanced contextual representation.
    \item We design a Multi-Scale Distribution-Aware (MSDA) deep supervision loss that preserves fine structural and boundary information across decoding stages.
    \item We conduct extensive experiments on seven public benchmarks, demonstrating that \textbf{Deco-Mamba} achieves SOTA performance with strong generalization and moderate model complexity.
\end{itemize}


\section{Methodology}
\label{sec:method}



\begin{figure*}[h]
\begin{center}
\includegraphics[width=0.75\linewidth, height=4.5in]{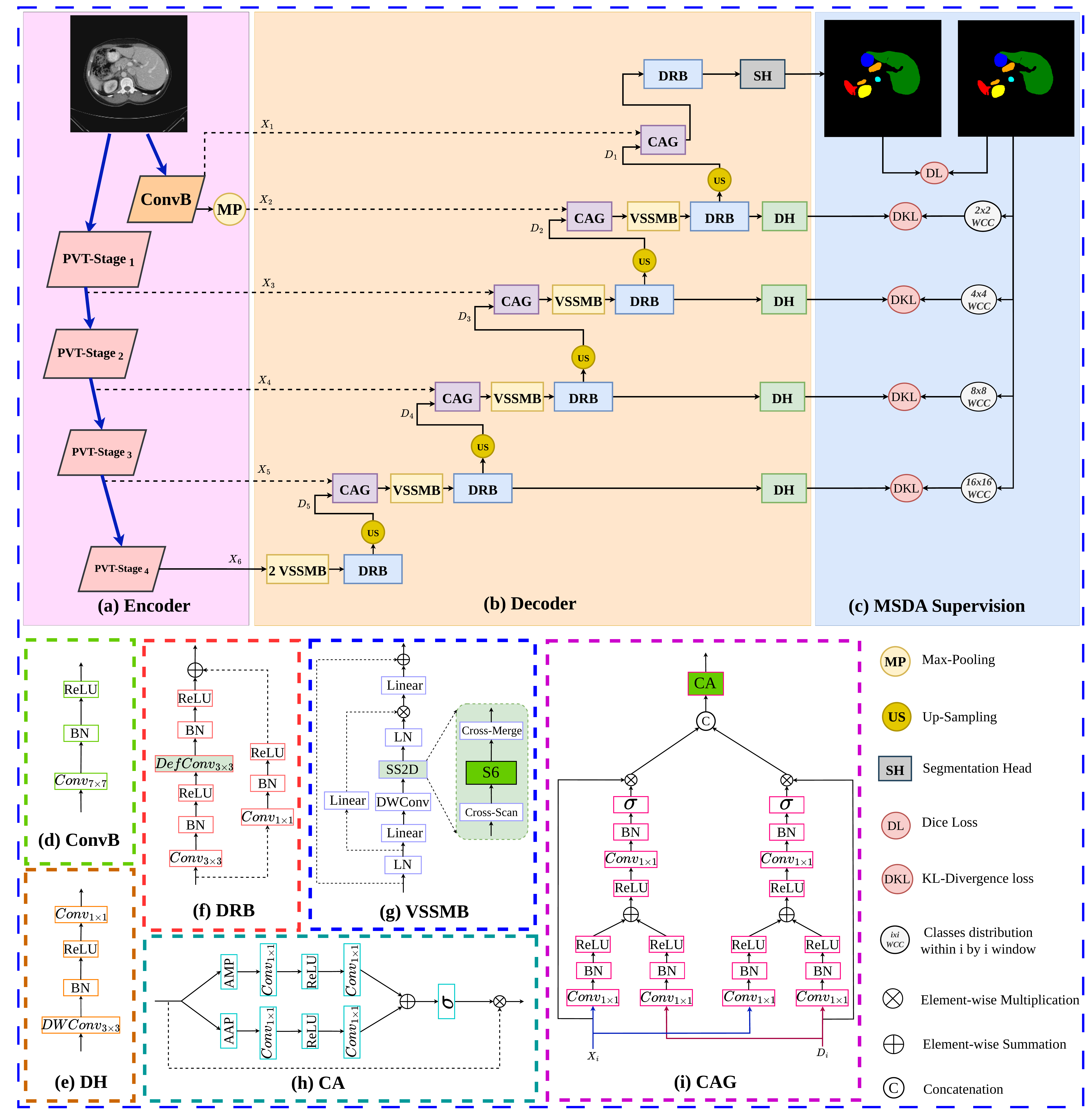} 
\caption{Our proposed Deco-Mamba architecture. (a) Encoder, (b) Decoder, (c) Multi-Scale Distribution-Aware Supervision, (d) Convolutional block, (e) Distribution Head, (f) Double Deformable Residual Block, (g) Visual-State Space Mamba Block, (h) Channel Attention Block, (i) Co-Attention Gate.}
\label{fig:approach}
\end{center}
\end{figure*}

This paper introduces \textbf{Deco-Mamba}, a hybrid encoder-decoder architecture that achieves strong global reasoning while preserving precise spatial detail. The encoder jointly extracts local CNN features and global Transformer dependencies, followed by a novel \textbf{CNN-Mamba decoder} designed for progressive global-to-local reconstruction with distribution-based deep supervision, as summarized in Figure \ref{fig:approach}.

\subsection{Encoder} 
Unlike prior segmentation frameworks that rely on heavy backbones, we deliberately adopt a \textbf{hybrid encoder} that captures both local and global representations at multiple scales. The input image $I \in \mathbb{R}^{H \times W \times C}$ is processed through a CNN branch that preserves high-resolution spatial details for fine-grained structures, while the PVT Transformer \cite{wang2022pvt} captures scalable long-range context at deeper encoder stages.

\paragraph{CNN branch.} As shown in Figure~\ref{fig:approach}.d, a $7\times7$ convolution (stride 1, padding 3), followed by batch normalization and ReLU activation function, extracts broad local context without reducing resolution: 

\begin{equation} 
X_1 = ReLU(BN(\text{Conv}_{7\times7}(I))), \quad X_2 = \text{MP}(X_1). 
\end{equation} 

These features encode fine-scale spatial detail critical for mask reconstruction. We forward $X_1$ and $X_2$ only to the last two decoder stages, directly reinforcing pixel-level precision during reconstruction.

\paragraph{Transformer branch.} In parallel, we employ a four-stage PVT hierarchy that progressively encodes global dependencies with attention of linear complexity in spatial resolution. Its outputs are denoted $X_3, X_4, X_5,$ and $X_6$, where $X_6$ is the bottleneck. This dual-branch encoder thus provides stronger inductive bias with lower complexity.
\subsection{Decoder} In contrast to most state-of-the-art methods that focus primarily on the encoding phase, with strong backbones that are usually computationally expensive and coupled with lightweight decoders, our architecture mainly focus on designing efficient decoder. Thus, we introduce a new hybrid \textbf{CNN--Mamba decoder}. 
As illustrated in Figure~\ref{fig:approach}.b, the proposed decoder comprises six stages. The bottleneck features from the encoder ($X_6$) are first passed to a decoder block consisting of two VSSMB modules, followed by a DDConv layer and an upsampling operation. The remaining decoder stages include the proposed \textbf{Co-Attention Gate}, VSSMB, and DDConv blocks, except for the final stage, which contains a CAG and DDConv followed by a segmentation head that maps the output channels to the number of target classes (N).

\noindent\textbf{Co-Attention Gate (CAG).} The \textit{Attention Gate (AG)}, originally introduced in \textit{Attention U-Net (Att-UNet)}~\cite{oktay2018attention}, uses decoder features as gating signals to highlight salient regions in the corresponding encoder features. For the $i$-th decoder stage, the AG is defined as:

\begin{equation} 
\begin{aligned} 
D_i' &= AG(x = X_i, g = D_{i+1}) \\ 
&= X_i \odot \sigma \big( \mathrm{Conv}_{1\times1}\big( \mathrm{Conv}_{1\times1}(X_i) + \mathrm{Conv}_{1\times1}(D_{i+1}) \big) \big) 
\end{aligned} 
\label{eq:2} 
\end{equation}

\noindent where $X_i$, $D_{i+1}$, and $\sigma$ denote the skip connection features from the encoder, the features from the previous decoder stage, and the sigmoid function, respectively. Despite its effectiveness in medical image segmentation, AG only emphasizes salient regions in encoder features and does not explicitly explore spatial saliency in decoder features. Moreover, it models spatial importance while ignoring channel-wise relationships.

To address these limitations, we propose the \textbf{Co-Attention Gate} (Figure~\ref{fig:approach}.i), which uses both encoder and decoder features as input and gating signals interchangeably. The outputs of the two attention gates are concatenated and refined by a \textbf{Channel Attention (CA)} (Figure~\ref{fig:approach}.h) block to emphasize the most informative channels. Formally, the CAG is defined as:

\begin{equation} \begin{aligned} D_i' &= CA[AG(x = X_i, g = D_{i+1}),AG(x = D_{i+1}, g = X_i)] \end{aligned} \label{eq:3} \end{equation}

\noindent where the \textbf{CA} operation is given by:

\begin{equation} 
\begin{aligned} 
CA(x) &= x \odot \sigma \Big( \mathrm{Conv}_{1\times1}\big( \mathrm{ReLU}(\mathrm{Conv}_{1\times1}(\mathrm{AMP}(x))) \big) \\ &\quad + \mathrm{Conv}_{1\times1}\big( \mathrm{ReLU}(\mathrm{Conv}_{1\times1}(\mathrm{AAP}(x))) \big) \Big) \end{aligned} 
\label{eq:4} 
\end{equation}

\noindent where $\mathrm{AMP}$ and $\mathrm{AAP}$ denote adaptive max pooling and adaptive average pooling, respectively.

\noindent\textbf{Mamba Block.} To efficiently capture long-range dependencies with linear computational complexity, we adopt a State-Space Model formulation as the core contextual modeling unit of our decoder. The proposed Visual State-Space Mamba Block (VSSMB) leverages a continuous-time SSM to model spatial correlations across both height and width dimensions. As illustrated in Figure~\ref{fig:approach}.g, each VSSMB integrates a lightweight convolutional layer for local feature enrichment, followed by a selective scan mechanism that propagates contextual information along multiple spatial directions (horizontal, vertical, and their inverses). This directional scanning enables effective global context modeling without relying on computationally expensive self-attention mechanisms.

Within the decoder, the VSSMB is positioned between the Co-Attention Gate (CAG) and the DDConv, refining the fused encoder–decoder features and reinforcing global contextual coherence before local refinement. This design ensures consistent semantic propagation while preserving spatial precision crucial for boundary delineation. Two VSSMB blocks are applied at the bottleneck to capture deep semantics at low spatial resolution $(H/32, W/32)$, followed by one block per stage from the second to fifth decoders for progressive refinement. The final stage omits the VSSMB, where convolutional operations better handle fine-grained reconstruction at full resolution $(H, W)$.

\noindent\textbf{Deformable Residual Block (DRB).}
During decoding, the DRB refines spatial representations and recovers anatomical details that may be smoothed by the global modeling of the VSSMB. While the VSSMB captures long-range dependencies and structured context, it may overlook subtle local variations crucial for precise segmentation. The DRB addresses this by introducing spatial adaptivity through deformable convolutions, enabling better modeling of geometric variations and complex tissue boundaries.

As illustrated in Figure~\ref{fig:approach}.f, each DRB integrates a standard $3\times3$ convolution and a deformable convolution within a residual framework, each followed by batch normalization and ReLU activation. The deformable convolution predicts pixel-wise offsets and modulation masks: the offset branch estimates sampling displacements, and the modulation branch assigns pixel-wise importance weights constrained to $[0,2]$ via a sigmoid activation. Positioned after each VSSMB, the DRB enhances boundary precision and fine-grained spatial details while maintaining global coherence. Together, VSSMB and DRB unify global context aggregation with local refinement for anatomically consistent segmentation.

\subsection{Multi-Scale Distribution-Aware Supervision (MSDA Supervision)}
To enhance both regional consistency and boundary precision, we introduce a \textbf{Multi-Scale Distribution-Aware Supervision (MSDA)} strategy that jointly enforces pixel-level accuracy and region-level statistical alignment. The supervision combines two complementary objectives: a Dice loss applied to the final decoder output and a distributional Kullback–Leibler (KL) divergence loss applied to intermediate decoder scales.
\paragraph{Dice Loss for Final Prediction.}
Let $\hat{Y} \in \mathbb{R}^{H \times W \times N}$ denote the predicted segmentation map and 
$Y \in \{0,1\}^{H \times W\times N}$ the corresponding ground truth. 
The Dice loss is defined as:
\begin{equation}
\mathcal{L}_{\text{dice}} = 1 - 
\frac{2 \sum_{h,w,n} \hat{Y}_{h,w,n} \, Y_{h,w,n}}
{\sum_{h,w,n} \hat{Y}_{h,w,n}^2 + \sum_{h,w,n} Y_{h,w,n}^2 + \epsilon},
\end{equation}
which promotes accurate spatial overlap and robust class balance between predictions and ground truth.

\paragraph{Deep Supervision with Distributional KL-Divergence Loss.}
Deep supervision at multiple decoder scales has been widely explored in medical image segmentation~\cite{zhou_unet_2018, fu2018joint, rahman2024multi, rahman2024emcad, linguraru_swin-umamba_2024}. 
However, most existing approaches resize intermediate outputs to match the ground truth resolution, leading to information loss and degradation of fine structural details. To address these limitations, we propose a distribution-aware multi-scale deep supervision scheme based on a KL-divergence loss that directly operates at the native resolution of each decoder scale.

For each intermediate decoder output 
$\hat{Y}^{(s)} \in \mathbb{R}^{H_s \times W_s \times C_s}$ at scale $s$, 
a \textbf{distribution head} projects the feature channels $C_s$ to the number of semantic classes $N$ as follow:

\begin{equation}
\hat{Z}^{(s)} = 
\text{Conv}_{1\times1}\!\left(
\text{ReLU}\!\left(
\text{BN}\!\left(
\text{DWConv}_{3\times3}(\hat{Y}^{(s)})
\right)\right)\right),
\end{equation}

\noindent where $\text{DWConv}_{3\times3}$ denotes a depthwise $3\times3$ convolution, followed by batch normalization (BN), ReLU activation, and a $1\times1$ convolution that produces $N$ output channels.  

To match the output resolution, the ground-truth distribution 
$\tilde{P}^{(s)} \in [0,1]^{H_s \times W_s \times N}$  is computed by averaging class frequencies within local spatial windows of size $K_h \times K_w$:
\begin{equation}
\tilde{P}^{(s)}_{b,c,h,w} = 
\frac{1}{K_h K_w} 
\sum_{i,j \in \Omega_{h,w}} \mathbb{I}(Y_{b,i,j} = c),
\end{equation}
where $\Omega_{h,w}$ denotes the set of pixels within the window centered at $(h,w)$.

The model’s predictions are converted to log-probabilities:
\begin{equation}
\log Q^{(s)} = \log \text{softmax}(\hat{Z}^{(s)}),
\end{equation}
and the distributional discrepancy is quantified using the KL divergence:
\begin{equation}
\mathcal{L}_{\text{KL}}^{(s)} = 
\sum_{b,h,w} \sum_{c=1}^{N} 
\tilde{P}^{(s)}_{b,c,h,w} 
\log \frac{\tilde{P}^{(s)}_{b,c,h,w}}{Q^{(s)}_{b,c,h,w}}.
\end{equation}

\paragraph{Boundary-Aware Weighting.}
To emphasize class transitions and ambiguous regions, 
we compute a \textbf{boundary weighting map} based on local class diversity:

\begin{equation}
W^{(s)}_{h,w} = 
\big(1 - \max_n \tilde{P}^{(s)}_{h,w,n}\big)^{\alpha}.
\end{equation}

\noindent Regions with mixed-class distributions (i.e., near object boundaries) receive higher weights.  The weighted distributional loss is thus formulated as:

\begin{equation}
\mathcal{L}_{\text{dist}}^{(s)} = 
\frac{1}{H_s W_s} 
\sum_{h,w} 
\big(1 + W^{(s)}_{h,w}\big) 
\, \mathcal{L}_{\text{KL}}^{(s)}(h,w).
\end{equation}

\paragraph{Multi-Scale Aggregation.}
Losses from multiple decoder scales are aggregated hierarchically, with progressively larger weights assigned to deeper decoder stages:

\begin{equation}
\mathcal{L}_{\text{multi}} = 
\sum_{s=1}^{S} \lambda_s \, \mathcal{L}_{\text{dist}}^{(s)},
\quad \text{where} \quad 
\lambda_1 < \lambda_2 < \dots < \lambda_S.
\end{equation}

The final objective function combines the pixel-level Dice loss and the multi-scale distributional supervision as:

\begin{equation}
\mathcal{L}_{\text{total}} = 
\mathcal{L}_{\text{dice}} + \mathcal{L}_{\text{multi}}.
\end{equation}

This formulation enforces both local proportion consistency and global structural alignment across scales. 
The Dice term ensures precise mask boundaries, while the distributional KL-divergence enhances regional coherence and boundary sensitivity.

\subsection{Deco-Mamba Variants}
We develop two variants of the proposed Deco-Mamba architecture based on different PVT backbones, namely \textbf{Deco-Mamba-V\textsubscript{0}} and \textbf{Deco-Mamba-V\textsubscript{1}}, which employ PVT-V2-B0 and PVT-V2-B2, respectively. Additional comparisons with alternative CNN and Transformer backbones are presented in the results section.

\begin{table}
\caption{Comparison of Abdominal Multi-Organ Segmentation on the Synapse Dataset. DSC and HD95 represent the average Dice Score and 95\% Hausdorff Distance across eight classes, respectively. Columns 4 to 11 present the Dice Score for each class, while the last two columns indicate model complexity in terms of the number of parameters (millions) and the number of FLOPs (GMac), respectively. All methods follow the same splitting and evaluation protocol introduced in \cite{chen2021transunet} for a fair comparison. Red and blue values indicate the best and second-best results, respectively.}
\begin{center}
\label{tab:synapse}
\centering
\resizebox{1\columnwidth}{!}{
\begin{tabular}{|l||c|c||c|c|c|c|c|c|c|c||c|c|}

\hline
{\multirow{2}{*} \textbf{Architecture}}  & \multicolumn{2}{|c|}{\textbf{Average}}& \multirow{2}{*} \textbf{Aorta}&\multirow{2}{*} \textbf{Gallbladder} &\multirow{2}{*} \textbf{Kidney (L) } &\multirow{2}{*} \textbf{Kidney (R)}& \multirow{2}{*} \textbf{Liver}& \multirow{2}{*} \textbf{Pancreas}& \multirow{2}{*} \textbf{Spleen}& \multirow{2}{*} \textbf{Stomach} & \multicolumn{2}{|c|}{\textbf{Complexity}} \\
       
\cline{2-3} \cline{12-13}
& \textbf{DSC$\uparrow$}& \textbf{HD95$\downarrow$} & &   & &  & &   & &  & \textbf{\#Params}& \textbf{\#FLOPs} \\
\hline\hline 

UNet \cite{ronneberger2015u}&  74.68 &36.87& 84.18& 62.84 &79.19& 71.29& 93.35 &48.23 &84.41 &73.92 & 7.85 & 10.73 \\\hline

Att-UNet \cite{oktay2018attention}&75.57& 36.97& 55.92& 63.91& 79.20& 72.71 &93.56& 49.37& 87.19 &74.95 & 7.98 & 11.05 \\\hline 

Unet++ \cite{zhou_unet_2018}  & 77.30 & 30.74 & 87.08 &  65.53 & 82.26 &  76.02 &93.98  & 53.61 & 86.09  & 73.85 & 9.16  &  26.52  \\\hline \hline

TransUNet \cite{chen2021transunet}& 77.48& 31.69 &87.23 &63.13 &81.87& 77.02 &94.08& 55.86 &85.08 &75.62 & 105.28 & 24.66  \\\hline

Swin-Unet \cite{liu_swin_2021} &77.58 &27.32 & 81.76 &65.95 &82.32 &79.22 &93.73& 53.81 &88.04 &75.79 & 41.38 &15.12  \\\hline

MTUnet \cite{wang2022mixed}& 78.59 & 26.59 &  87.92&   64.99& 81.47& 77.29 & 93.06& 59.46  &  87.75 & 76.81 & 79.07 & 44.73 \\\hline

UCTransNet \cite{wang2022uctransnet}& 78.23 & 26.75 & \bf{-} &  \bf{-} & \bf{-}& \bf{-} & \bf{-}&  \bf{-} &  \bf{-} & \bf{-} & 66.43 &32.94  \\\hline

TransClaw U-Net \cite{wang2022uctransnet}& 78.09 & 26.38 &85.87 & 61.38  & 84.83& 79.36 & 94.28&  57.65 &  87.74 &  73.55 & \bf{-} &  \bf{-} \\\hline
          
ST-Unet \cite{zhang2023st} &78.86 &20.37 &85.68& 69.05& 85.81&  73.04&  95.13&60.23&   89.15&  72.78 & \bf{-} &  \bf{-}
\\\hline

TransCeption \cite{azad2023enhancing}& 82.24& 20.89 &87.60 &71.82& 86.23 &80.29 &95.01 &65.27 &\textcolor{blue}{91.68} &80.02 & \bf{-} &  \bf{-}
\\\hline

PVT-EMCAD-B2 \cite{rahman2024emcad}& 82.66&  19.35 &85.61 & 67.28 &86.31 & 83.94  &95.22 & 67.28 &91.12 & \textcolor{blue}{84.51} & 26.77 & 4.44  \\\hline

Parallel-MERIT \cite{rahman2024multi}& 82.43&  16.20 &87.37 & 69.03 &85.35 & 82.48  &94.97 & 67.05 &90.18 & 83.02 & 147.85  & 34.19 \\\hline

Cascaded-MERIT \cite{rahman2024multi}& \textcolor{blue}{83.59}&  15.99  & 86.30 & \textcolor{red}{75.22} &  86.39&   83.27 & 94.71 &  67.55 & 91.15 &  84.12 & 147.86  & 33.31 \\\hline

PAG-TransYnet \cite{bougourzi2024rethinking}& 83.43&  \textcolor{blue}{15.82} &\textcolor{blue}{89.67} & 68.89 &86.74 & \textcolor{red}{84.88}  &\textcolor{blue}{95.87} &\textcolor{blue}{68.75} &\textcolor{red}{92.01} & 80.66  & 144.22 &33.65   \\\hline\hline

UMamba\_Bot\_2D \cite{U-Mamba}& 74.74&  29.26 &86.37 & 59.45 &81.50 & 72.29  &93.82 & 51.12 &88.20 & 65.14 & 21.33 & 11.26  \\\hline

UMamba\_Enc\_2D \cite{U-Mamba}& 76.30&  27.23 &87.08 & 59.36 &78.60 & 70.14  &93.63 & 58.14 &88.48 & 74.98 & 21.62 & 9.95   \\\hline

Mamba-Unet \cite{wang2024mamba}& 76.21&  22.64 &85.40 & 65.88 &84.51 & 76.65  &93.46 & 49.50 &85.68 & 68.61 & 19.12 & 4.53   \\\hline

HC-MAMBA \cite{xu2024hc} &79.58 &26.34& 89.93& 67.65& 84.57& 78.27 &95.38& 52.08& 89.49& 79.84&  12  &   -
\\ \hline 

Swin-UMamba \cite{linguraru_swin-umamba_2024}& 80.34&  21.51 &87.46 & 66.45 &84.39 & 78.49  &95.17 & 63.35 &90.08 & 77.30 & 59.88 & 31.35   \\\hline

VM-UNet  \cite{ruan2024vm} & 81.08 & 19.21 &  86.40& 69.41 &86.16 &82.76& 94.17& 58.80& 89.51& 81.40& 27.43  &   4.11 \\ \hline

HCMUNet \cite{ma2025u}& 81.52& 17.83& 88.06& 69.60& 87.04& 82.35& 95.10& 59.24& 90.63& 80.76 &- &- \\ \hline

SliceMamba \cite{fan2025slicemamba}& 81.95& 16.04  &87.78  &68.77 &\textcolor{blue}{88.30} &\textcolor{blue}{84.26}  &95.25 &64.49  &86.91 &79.82  & - &  -  \\\hline

 \hline 
Deco-Mamba-V\textsubscript{0} & 83.16& 15.89 &\textcolor{red}{89.98} & 68.62 & 88.15& 83.01  & 95.68& 64.07 &91.54 & 84.28 &  9.67&  9.73
\\\hline

\textbf{Deco-Mamba-V\textsubscript{1}}& \textcolor{red}{85.07}& \textcolor{red}{14.72}&89.62& \textcolor{blue}{72.53} & \textcolor{red}{88.76}& 84.18 &\textcolor{red}{95.94} & \textcolor{red}{72.02}  &91.60 & \textcolor{red}{85.94} &46.93 & 17.24

\\\hline
\end{tabular}}
\end{center}
\end{table}

\begin{table}
\caption{Comparison of abdominal multi-organ segmentation on the BTCV dataset (13 classes). \textbf{DSC} and \textbf{HD95} denote the average Dice score and 95\% Hausdorff distance over all classes, respectively. Columns four to the last report the Dice score for each organ. All methods follow the same data split and evaluation protocol as in \cite{chen2021transunet}. Red and blue values indicate the best and second-best results, respectively.}
\begin{center}
\label{tab:BTCV}
\centering
\resizebox{1\columnwidth}{!}{
\begin{tabular}{|l||c|c||c|c|c|c|c|c|c|c|c|c|c|c|c|}
\hline
\multirow{2}{*}{Architecture} & \multicolumn{2}{c|}{\textbf{Average}}       & \multirow{2}{*}{Spleen} & \multirow{2}{*}{Kidney (R)} & \multirow{2}{*}{Kidney (L)} & \multirow{2}{*}{Gallbladder} & \multirow{2}{*}{Esophagus} & \multirow{2}{*}{Liver} & \multirow{2}{*}{Stomach} & \multirow{2}{*}{Aorta} & \multirow{2}{*}{IVA} & \multirow{2}{*}{P\&S Vein} & \multirow{2}{*}{Pancreas} & \multirow{2}{*}{AG (R)} & \multirow{2}{*}{AG (L)} \\ \cline{2-3}
 & \multicolumn{1}{l|}{\textbf{DSC$\uparrow$}}   & \textbf{HD95$\downarrow$}    &    &        &      &    &    &      &       &     &    &       &       &    &      \\ \hline
U-Net \cite{ronneberger2015u}&71.35 &25.16 & 84.71 &75.87     & 82.05 & 63.38 & 69.23 &92.71  & 71.66 & 86.57 & 71.93 & 60.82 & 54.85 & 57.14   &56.58  \\ \hline

Att-UNet \cite{oktay2018attention}&71.75 &26.16 & 84.21 & 77.40    &  82.60& 65.91 & 68.72 & 93.53 & 73.89 & 86.12 &  70.86& 61.52 &54.10  & 55.29   &  58.63\\ \hline

Unet++ \cite{zhou_unet_2018}& 72.18&21.53 & 85.70 &  76.54   & 84.28 &67.23 & 72.18 & 93.78 & 68.51 & 87.08 & 72.01 &61.91  & 54.13& 60.13  &  54.79\\ \hline\hline

Swin-Unet \cite{liu_swin_2021} &67.65 &21.52 & 85.48&73.32 &80.05 &62.76 &65.26& 92.53 &71.15 & 79.03& 63.45 &54.26 &54.81  &51.12 &46.26 \\\hline 

TransUNet \cite{chen2021transunet} & 71.85  & 22.02 & 85.31  & 79.26 &83.38  & 61.68 &67.54  &94.03  & 74.73  & 87.04 &75.75  & 59.91 & 58.18  &57.96  & 49.28 \\\hline

UCTransNet \cite{UCTransNet}  & 72.45 & 24.38 & 86.09  & 70.38   & 80.62  & 63.47  & 74.27  & 93.41  & 74.17  & 86.52   & 76.81  & 62.27 & 55.00  & \textcolor{blue}{61.27} & 57.60    \\ \hline
PVT-EMCAD-B2 \cite{rahman2024emcad}  & 72.41 & 14.52 & 88.95    & 79.27 & 83.92  & 66.12  & 71.44   & 95.50  & 81.49   & 86.06   & 75.67  & 64.71           & 64.09  & 45.95   & 38.12   \\ \hline 
Cascaded-MERIT \cite{rahman2024multi}  & 72.74 & 16.10 & 89.45 & 76.38  & 80.36 & 68.03 & 70.03  & 94.37  & 79.60   & 85.12 & 71.88  & 65.79  & 61.72  & 53.01    & 49.88    \\ \hline
Parallel-MERIT \cite{rahman2024multi}  & 74.78 & 13.53 & \textcolor{red}{90.70} & \textcolor{blue}{83.00} & 86.17  & 68.30  & 68.42   & 94.60  & 82.70   & 85.48   & 74.83  & 68.57  & \textcolor{blue}{67.04}  & 50.91   & 51.40    \\ \hline

PAG-TransYnet \cite{bougourzi2024rethinking}  & 75.87 & 17.02 & \textcolor{blue}{90.08} & 78.34 & 80.42  &  65.68 & 72.99   & \textcolor{red}{95.71}  & \textcolor{blue}{83.63}   & \textcolor{blue}{88.59}  &78.27  & \textcolor{blue}{68.94}  & 64.97  & 60.31  & \textcolor{blue}{58.36}   \\ \hline

\hline \hline
VM-UNet  \cite{ruan2024vm} &62.79 & 18.34 &87.74  & 74.61 & 81.55   & 59.34   & 55.62 & 93.23  & 63.57 & 78.26 & 59.87& 51.34 &42.68 & 36.13& 32.26 \\ \hline

UMamba\_Bot\_2D \cite{U-Mamba}  & 67.30 & 25.23 & 84.40     & 71.05  & 79.22  & 47.30   & 63.82  & 93.43 & 69.07   & 86.07  & 69.32    & 53.54  & 56.72   & 51.19    & 49.82    \\ \hline
UMamba\_Enc\_2D \cite{U-Mamba}  & 65.97 & 14.97 & 87.37     & 69.82  & 78.25  & 59.55   & 67.27  & 94.05   & 72.81   & 85.31  & 72.37  & 58.96  & 59.48   & 30.39    & 22.00    \\ \hline 
Mamba-Unet \cite{wang2024mamba}  &69.92 &24.10  &  84.03    &72.96  &78.88   & 65.22 &  68.95 & 94.14  & 74.17   &  83.21  &73.33 & 57.04  &56.77  & 52.98  &   47.26     \\ \hline
Swin-UMamba \cite{linguraru_swin-umamba_2024}  &73.40  &16.22  & 85.84  & 81.58 & 84.73 & 59.43  &  69.42  &94.21  &76.58    & 85.53   &76.66   &  68.34  & 61.72  &  55.31  &  54.87   \\ \hline\hline

Deco-Mamba-V\textsubscript{0} &\textcolor{blue}{76.29}&\textcolor{blue}{12.02}&89.59 &82.54  &\textcolor{red}{88.98} & \textcolor{blue}{69.31} &\textcolor{blue}{74.66} &  \textcolor{blue}{95.67} & 76.96& 88.02 &\textcolor{blue}{80.64} & 64.59 & 65.28 & 59.87&55.64
\\\hline

Deco-Mamba-V\textsubscript{1}& \textcolor{red}{78.45} &\textcolor{red}{11.77}&88.88  & \textcolor{red}{83.10}  &\textcolor{blue}{86.94} & \textcolor{red}{69.68}  & \textcolor{red}{75.58}    & 95.51 &  \textcolor{red}{84.33} &  \textcolor{red}{88.72} &  \textcolor{red}{81.40} & \textcolor{red}{71.47}   &   \textcolor{red}{70.84} & \textcolor{red}{61.42}  &  \textcolor{red}{61.93}  
\\\hline

\end{tabular}}
\end{center}
\end{table}

\section{Experiments and Results}

\subsection{Datasets}
Our approach is evaluated using seven benchmark datasets: Synapse \cite{landman2015miccai}, BTCV \cite{landman2015miccai}, ACDC \cite{bernard2018deep},  ISIC17 \cite{codella2018skin}, ISIC18 \cite{codella2019skin}, GlaS \cite{sirinukunwattana2017gland}, and MoNuSeg \cite{kumar2017dataset}. These datasets include variety of imaging modalities and tasks. For abdominal multi-organs segmentation from CT-scans, Synapse and BTCV, we followed the same splitting and evaluation protocol introduced in \cite{chen2021transunet} for segmenting 9 and 13 classes, respectively. For Automated Cardiac Diagnosis Challenge (ACDC) dataset, we followed the same splitting introduced in \cite{wang2022mixed}. For ISIC17 and ISIC18 datasets, we followed the official splitting \cite{codella2018skin, codella2019skin}. For GlaS and MoNuSeg, we followed the evaluation protocol introduced in \cite{UCTransNet}. More details about the datasets, evaluation protocol and metrics are available in the Supplementary Materials (Sections A.1).

\subsection{Implementation settings}
We implemented our model in PyTorch and trained it on an NVIDIA A5000 GPU (24~GB). Standard augmentations, including random horizontal/vertical flipping and rotation, were applied. We used the AdamW optimizer with a cosine learning rate schedule and warm restarts ($T{=}2$). All datasets were trained with $224{\times}224$ resolution inputs. 
Unless stated otherwise, the learning rate and batch size were $1{\times}10^{-4}$ and $16$, respectively; for GlaS and MoNuSeg, $3{\times}10^{-4}$ and $4$ were used. Models were trained for 120~epochs on Synapse and BTCV, 100~epochs on ISIC17, ISIC18, GlaS and MoNuSeg, and 400~epochs on ACDC, with the best validation model (or final model for Synapse/BTCV) used for testing.

\subsection{Results}

Tables~\ref{tab:synapse}--\ref{tab:binary} summarize the evaluation results and comparisons with state-of-the-art methods across seven benchmark datasets. The compared approaches include baseline CNNs, Transformer-based models, and recent Mamba-based architectures. 

For abdominal multi-organ segmentation on the Synapse dataset (Table~\ref{tab:synapse}), the proposed \textbf{Deco-Mamba-V\textsubscript{1}} achieves the best performance, surpassing the second-best method (Cascaded-MERIT) by 1.48\% in Dice and 1.27 in HD95. Meanwhile, \textbf{Deco-Mamba-V\textsubscript{0}} attains comparable performance to the strongest Transformer-based models while requiring approximately 15$\times$ fewer parameters. Compared to other Mamba-based approaches, Deco-Mamba-V\textsubscript{0} and Deco-Mamba-V\textsubscript{1} outperform SliceMamba by 1.21\% and 3.12\% in Dice, respectively. In per-organ analysis, Deco-Mamba-V\textsubscript{1} ranks first in half of the classes and second in a quarter, while the next best competitors (Cascaded-MERIT and PAG-TransYNet) are Transformer-based architectures with around 150M parameters.

Similar trends are observed on the BTCV dataset (Table~\ref{tab:BTCV}), where \textbf{Deco-Mamba-V\textsubscript{1}} and \textbf{Deco-Mamba-V\textsubscript{0}} achieve the best and second-best performances, respectively. Deco-Mamba-V\textsubscript{1} obtains the highest Dice score in ten out of thirteen organ classes. Although the strongest competitors remain Transformer-based models, they incur substantially higher computational costs. In contrast, most Mamba-based methods perform worse than baseline CNNs, except Swin-UMamba, which still requires roughly 6$\times$ more parameters than our lighter \textbf{Deco-Mamba-V\textsubscript{0}}. These results indicate that while most methods, especially Mamba-based ones, are sensitive to task complexity, our approach achieves a favorable balance between segmentation performance and computational efficiency.

\begin{table}
\caption{Comparison on Automatic Cardiac Diagnosis Challenge (ACDC) dataset. Red and blue values indicate the best and second-best results, respectively.   }
\begin{center}
\label{tab:ACDC}
\centering
\begin{tabular}{|l|l|c|c|c|c|}
\hline
Ex & Architecture   & \textbf{Avg DSC $\uparrow$} & RV  $\uparrow$  & Myo  $\uparrow$ & LV  $\uparrow$  \\ \hline

1  & U-Net \cite{ronneberger2015u}        &  87.84 &   86.51& 84.66 &92.36 \\ \hline
2  & Att-UNet  \cite{oktay2018attention}     & 88.04 &86.70 & 84.59 &  92.83  \\ \hline

3  & UNet++ \cite{zhou_unet_2018}& 90.65& 88.52&88.31 &  95.12    \\ \hline\hline

4  & SwinUNet \cite{cao2022swin}      & 88.07   & 85.77 & 84.42 & 94.03 \\ \hline

5  & TransUNet \cite{chen2021transunet}     & 89.71   & 86.67 & 87.27 & 95.18 \\ \hline

6  & MT-Unet \cite{wang2022mixed}       & 90.43   & 86.64 & 89.04 & 95.62 \\ \hline
7  & MISSFormer \cite{huang2022missformer}     & 90.86   & 89.55 & 88.04 & 94.99 \\ \hline

8& PAG-TransYnet \cite{bougourzi2024rethinking}&90.89  & 89.64& 88.48& 94.53\\\hline 

9&UCTransNet \cite{UCTransNet} & 91.25 &88.85 &89.28 & 95.60\\\hline 

10  & Cascaded-MERIT \cite{rahman2024multi} & 91.85   & 90.23 & 89.53 & 95.80 \\ \hline

11  & PVT-EMCAD-B2 \cite{rahman2024emcad}   & \textcolor{blue}{92.12}   & \textcolor{blue}{90.65} & 89.68 & \textcolor{blue}{96.02} \\ \hline\hline
12  &VM-UNet  \cite{ruan2024vm} & 88.61 & 85.33& 86.16& 94.33 \\\hline 
13  & Mamba-Unet \cite{wang2024mamba}   &  89.11  & 86.89 & 85.58 & 94.86 \\ \hline 

14  & Swin-UMamba \cite{linguraru_swin-umamba_2024}   & 91.18   & 89.34 & 88.94 & 95.26 \\ \hline
15 & HCMUNet \cite{ma2025u}    & 92.11   & \textcolor{red}{91.50}  &  \textcolor{blue}{90.20} & 94.61\\ \hline \hline
16  & \textbf{Deco-Mamba-V\textsubscript{0}}   &  91.55  & 89.77 & 89.34& 95.54 \\ \hline

17 & \bf{Deco-Mamba-V\textsubscript{1}}       & \textcolor{red}{92.35}   & 90.49 & \textcolor{red}{90.39} & \textcolor{red}{96.16} \\ \hline
\end{tabular}
\end{center}
\end{table}  
\begin{table}
\caption{Comparison on the ISIC17, ISIC18, GlaS, and MoNuSeg datasets. Red and blue indicate the best and second-best results, respectively. In our experiments, different metrics showed similar trends; only Dice scores are reported here, with additional metrics in the supplementary material (Tables~ 1 and 2 page 2).}
\begin{center}
\label{tab:binary}
\centering
\begin{tabular}{|c|l||c|c||c|c|}
\hline
Ex& \textbf{Architecture}& \textbf{ISIC17} & \textbf{ISIC18} 
& \textbf{GlaS } & \textbf{MoNuSeg} \\
\hline
1  & U-Net & 81.59 & 85.80 & 85.45 & 76.45 \\
\hline
2  & Att-UNet & 80.82 & 84.99 & 88.80 & 76.67 \\
\hline
3  & UNet++ & 80.64 & 85.59 & 87.56 & 77.01 \\
\hline \hline
4  & TransUNet & 84.10 & 86.49 & 88.40 & 78.53 \\
\hline
5  & FAT-Net & 85.00 & 86.91 & -- & -- \\
\hline
6  & PAG-TransYnet & 83.99 & 87.99 & 94.20 & 79.62 \\
\hline
7  & UCTransNet & 83.46 & 87.24 & 90.18 & 79.08 \\
\hline
8  & Swin-UNet & 85.01 & 89.04 & 89.58 & 77.69 \\
\hline
9  & Cascaded-MERIT & \textcolor{blue}{85.67} & 89.23 & \textcolor{red}{96.91} & 77.56 \\
\hline
10 & PVT-EMCAD-B2 & 84.33 & \textcolor{blue}{89.40} & 96.38 & 77.42 \\
\hline \hline
11 & Mamba-Unet & 84.56 & 87.04 & 93.41 & 76.54 \\
\hline
12 & Swin-UMamba & 85.47 & 87.08 & 95.85 & 81.45 \\
\hline
13 & VM-UNet & 81.67 & 88.32 & 94.25 & 77.29 \\
\hline \hline
14 & \textbf{Deco-Mamba-V\textsubscript{0}} 
& 85.14 & \textcolor{red}{89.67} 
& 96.61 & \textcolor{blue}{83.41} \\
\hline
15 & \textbf{Deco-Mamba-V\textsubscript{1}} 
& \textcolor{red}{86.01} & 89.36 
& \textcolor{red}{96.91} & \textcolor{red}{85.14} \\
\hline
\end{tabular}
\end{center}
\end{table}

As shown in Table~\ref{tab:ACDC}, \textbf{Deco-Mamba-V\textsubscript{1}} achieves state-of-the-art performance on the ACDC dataset, while \textbf{Deco-Mamba-V\textsubscript{0}} attains comparable accuracy to heavy Transformer-based models such as Cascaded-MERIT and UCTransNet. 
From Table~\ref{tab:binary}, Deco-Mamba-V\textsubscript{1} achieves the best overall performance, ranking first on ISIC17, GlaS (tied), and MoNuSeg, while Deco-Mamba-V\textsubscript{0} obtains the best result on ISIC18 and competitive second-best scores on GlaS and MoNuSeg. Notably, Deco-Mamba-V\textsubscript{1} improves Dice by +4.46\% over Swin-UMamba and +8.69\% over U-Net on MoNuSeg, and surpasses Transformer-based competitors such as Cascaded-MERIT and PVT-EMCAD-B2 across most datasets.

Overall, the proposed Deco-Mamba variants demonstrate strong cross-dataset generalization, consistently outperforming both CNN- and Transformer-based architectures while maintaining computational efficiency. Qualitative analysis is available in the Supplementary Materials (Section A.3)

\subsection{Ablation study}
In Table \ref{tab:ablation_all}, we analyze the contribution of each component in our decoder-centric architecture on the Synapse dataset. The first row quantifies the impact of the CNN branch on preserving fine-grained details, while the second row shows the effect of VSSMB on capturing long-range dependencies during the decoding phase.

Experiments 3–5 evaluate the efficiency of different attention gate mechanisms (AG \cite{oktay2018attention}, LGAG \cite{rahman2024emcad}, and CBAM \cite{woo2018cbam}). Results (rows 3–5) demonstrate the superiority of the proposed CAG. Although CBAM benefits from channel attention and outperforms other attention gates, it remains less effective than our CAG. Experiments 6–8 assess alternatives to the proposed DRB, showing that they cannot compensate for the loss of fine details caused by the SSM. The DRB is specifically designed to recover fine details while leveraging deformable offsets to focus on the most relevant spatial locations during reconstruction.

Regarding MSDA, experiments in rows 8–12 of Table \ref{tab:ablation_all} confirm the effectiveness of the proposed MSDA loss, whereas other boundary losses yield only marginal improvements. Conventional deep supervision (Experiment 10) slightly improves Dice scores compared to Experiment 9 but increases HD95, as early layers are forced to produce coarse predictions at lower resolutions, reducing boundary precision. In contrast, MSDA supervision improves both metrics by providing consistent, distribution-aware, multi-scale guidance during training.



\begin{table*}[t]
\caption{Ablation study on the Synapse dataset. We analyze encoder components, attention design, reconstruction strategies, and supervision losses under a unified setting.}
\label{tab:ablation_all}
\centering
\begin{tabular}{|c|l||c|c|}
\hline
\textbf{ID} & \textbf{Ablation Scenario} & \textbf{DSC$\uparrow$} & \textbf{HD95$\downarrow$}  \\
\hline\hline

1 & w/o CNN branch in encoder & 84.07 & 18.92  \\
2 & w/o VSSMB & 83.51& 15.96  \\
\hline\hline

3 & Replace CAG with AG \cite{oktay2018attention}& 82.98 & 15.69  \\
4 & Replace CAG with LGAG (EMCAD \cite{rahman2024emcad}) & 82.69 & 20.36  \\
5 & Replace CAG with CBAM \cite{woo2018cbam} & 84.01 & 16.19  \\
\hline\hline

6 & Replace DefConv with Standard Conv & 84.53 & 16.18 \\
7 & Replace DefConv with Involution \cite{li2021involution} & 83.92 & 20.57  \\
8 & Replace DefConv with Dynamic Conv \cite{chen2020dynamic} & 83.77 & 22.03  \\
\hline\hline

9 & Dice Loss only & 83.84 & 14.94 \\
10 & Dice + Deep Supervision& 84.24 &15.89  \\
11 & Dice + Boundary-aware Edge Loss \cite{takikawa2019gated}   & 83.94 & 21.43  \\
12 & Dice + Distance-based Boundary Loss \cite{kervadec2019boundary} & 84.12 & 20.64  \\
\hline\hline

13 & \textbf{Deco-Mamba (Dice + MSDA) – Ours} & \textbf{85.07} & \textbf{14.72}  \\
\hline

\end{tabular}
\end{table*}

From Table \ref{tab:back}, CNN-based backbones show lower computational complexity (notably in FLOPs) but inferior performance due to their limited ability to model long-range dependencies, while other Transformer backbones (Swin-T) can offer strong performance at higher computational cost.

\begin{table}
\caption{Ablation Study of Different Backbones}
\begin{center}
\label{tab:back}
\centering
\resizebox{0.6\columnwidth}{!}{
\begin{tabular}{|l||c|c||c|c|}
\hline
{\multirow{2}{*}{\textbf{En Backbone}}} & \multicolumn{2}{|c||}{\textbf{Average}} & \multicolumn{2}{|c|}{\textbf{Complexity}} \\
\cline{2-5}
& \textbf{DSC$\uparrow$} & \textbf{HD95$\downarrow$} & \textbf{\#Params} & \textbf{\#FLOPs} \\
\hline\hline
Efficientnet-B0 \cite{tan2019efficientnet}& 81.10 & 25.20 & 10.28 & 5.89 \\
\hline
RegNetX-400MF \cite{radosavovic2020designing}& 81.20 & 27.84 & 14.9 & 7.62 \\
\hline
Swin-T \cite{liu2021swin}& 83.76 & 14.30 & 70.12 & 22.66 \\
\hline
PVT-V2-B0 \cite{wang2022pvt}& 83.16 & 15.89 & 9.67 & 9.73 \\
\hline
PVT-V2-B2 \cite{wang2022pvt}& 85.07 & 14.72 & 46.93 & 17.24 \\
\hline
\end{tabular}}
\end{center}
\end{table}

\textbf{On Decoder-Centric Design:} \textit{Deco-Mamba} focuses on decoder-centric design while reducing dependency on large pretrained encoders. Compared  to the strongest competitors on 28-class segmentation (Fig. \ref{fig:params}), \textit{Deco-Mamba-V\textsubscript{0}}  achieves superior performance while relying on significantly lighter pretrained backbones (3.4M params / 0.6 GFLOPs) versus Swin-UMamba (30M / 4.9G), PAG-TransUNet (107.9M / 20.4G), Cascaded-MERIT (130.5M / 24.6G), and PVT-EMCAD-B2 (25.4M / 4G). Moreover, when using identical backbones, Deco-Mamba consistently outperforms prior methods: e.g., on Synapse, with Swin-T, our method achieves 83.76\% DSC (Table \ref{tab:back}) vs. 77.58\% for Swin-UNet (Table \ref{tab:synapse}). Similarly, with PVT backbones, Deco-Mamba using PVT-B0 surpasses EMCAD using PVT-B2 (Fig. \ref{fig:params}). These results indicate that performance gains stem primarily from the proposed decoder design rather than reliance on larger pretrained backbones.

\section{Conclusion}
\label{sec:conclusion}
In this work, we introduced Deco-Mamba, an efficient and generalizable framework for medical image segmentation that emphasizes the often-overlooked role of the decoder. By integrating Co-Attention Gates, Vision State Space Modules, and deformable convolutions within a CNN–Transformer–Mamba design, our model effectively captures both global context and fine structural details. Furthermore, the proposed multi-scale distribution-aware deep supervision provides stable and precise gradient guidance, improving boundary delineation and overall segmentation quality. Extensive experiments across diverse benchmarks demonstrate that Deco-Mamba achieves state-of-the-art performance with moderate computational cost, highlighting the importance of decoder design and distribution-aware supervision in advancing medical image segmentation.


%
%

\begin{thebibliography}{10}
\providecommand{\url}[1]{\texttt{#1}}
\providecommand{\urlprefix}{URL }
\providecommand{\doi}[1]{https://doi.org/#1}

\bibitem{azad2023enhancing}
Azad, R., Jia, Y., Aghdam, E.K., Cohen-Adad, J., Merhof, D.: Enhancing medical image segmentation with transception: a multi-scale feature fusion approach. arXiv preprint arXiv:2301.10847  (2023)

\bibitem{bernard2018deep}
Bernard, O., Lalande, A., Zotti, C., Cervenansky, F., Yang, X., Heng, P.A., Cetin, I., Lekadir, K., Camara, O., Ballester, M.A.G., et~al.: Deep learning techniques for automatic mri cardiac multi-structures segmentation and diagnosis: is the problem solved? IEEE transactions on medical imaging  \textbf{37}(11),  2514--2525 (2018)

\bibitem{bougourzi2023pdatt}
Bougourzi, F., Distante, C., Dornaika, F., Taleb-Ahmed, A.: {PDA}tt-{U}net: Pyramid dual-decoder attention unet for covid-19 infection segmentation from ct-scans. Medical Image Analysis  \textbf{86},  102797 (2023)

\bibitem{bougourzi2024rethinking}
Bougourzi, F., Dornaika, F., Taleb-Ahmed, A., Hoang, V.T.: Rethinking attention gated with hybrid dual pyramid transformer-cnn for generalized segmentation in medical imaging. International Conference on Pattern Recognition (ICPR)  (2024)

\bibitem{bougourzi2025recent}
Bougourzi, F., Hadid, A.: Recent advances in medical imaging segmentation: A survey. arXiv preprint arXiv:2505.09274  (2025)

\bibitem{cao2022swin}
Cao, H., Wang, Y., Chen, J., Jiang, D., Zhang, X., Tian, Q., Wang, M.: Swin-unet: Unet-like pure transformer for medical image segmentation. In: European conference on computer vision. pp. 205--218. Springer (2022)

\bibitem{chen2021transunet}
Chen, J., Lu, Y., Yu, Q., Luo, X., Adeli, E., Wang, Y., Lu, L., Yuille, A.L., Zhou, Y.: Transunet: Transformers make strong encoders for medical image segmentation. arXiv preprint arXiv:2102.04306  (2021)

\bibitem{chen2020dynamic}
Chen, Y., Dai, X., Liu, M., Chen, D., Yuan, L., Liu, Z.: Dynamic convolution: Attention over convolution kernels. In: Proceedings of the IEEE/CVF conference on computer vision and pattern recognition. pp. 11030--11039 (2020)

\bibitem{codella2019skin}
Codella, N., Rotemberg, V., Tschandl, P., Celebi, M.E., Dusza, S., Gutman, D., Helba, B., Kalloo, A., Liopyris, K., Marchetti, M., et~al.: Skin lesion analysis toward melanoma detection 2018: A challenge hosted by the international skin imaging collaboration (isic). arXiv preprint arXiv:1902.03368  (2019)

\bibitem{codella2018skin}
Codella, N.C., Gutman, D., Celebi, M.E., Helba, B., Marchetti, M.A., Dusza, S.W., Kalloo, A., Liopyris, K., Mishra, N., Kittler, H., et~al.: Skin lesion analysis toward melanoma detection: A challenge at the 2017 international symposium on biomedical imaging (isbi), hosted by the international skin imaging collaboration (isic). In: 2018 IEEE 15th international symposium on biomedical imaging (ISBI 2018). pp. 168--172. IEEE (2018)

\bibitem{dosovitskiy2020vit}
Dosovitskiy, A., Beyer, L., Kolesnikov, A., Weissenborn, D., Zhai, X., Unterthiner, T., Dehghani, M., Minderer, M., Heigold, G., Gelly, S., Uszkoreit, J., Houlsby, N.: An image is worth 16x16 words: Transformers for image recognition at scale. ICLR  (2021)

\bibitem{fan2025slicemamba}
Fan, C., Yu, H., Huang, Y., Wang, L., Yang, Z., Jia, X.: Slicemamba with neural architecture search for medical image segmentation. IEEE Journal of Biomedical and Health Informatics  (2025)

\bibitem{fu2018joint}
Fu, H., Cheng, J., Xu, Y., Wong, D.W.K., Liu, J., Cao, X.: Joint optic disc and cup segmentation based on multi-label deep network and polar transformation. IEEE transactions on medical imaging  \textbf{37}(7),  1597--1605 (2018)

\bibitem{gu2024mamba}
Gu, A., Dao, T.: Mamba: Linear-time sequence modeling with selective state spaces. In: First conference on language modeling (2024)

\bibitem{gu2021efficiently}
Gu, A., Goel, K., R{\'e}, C.: Efficiently modeling long sequences with structured state spaces. In: The International Conference on Learning Representations (ICLR) (2022)

\bibitem{huang2022missformer}
Huang, X., Deng, Z., Li, D., Yuan, X., Fu, Y.: Missformer: An effective transformer for 2d medical image segmentation. IEEE Transactions on Medical Imaging  (2022)

\bibitem{kervadec2019boundary}
Kervadec, H., Bouchtiba, J., Desrosiers, C., Granger, E., Dolz, J., Ayed, I.B.: Boundary loss for highly unbalanced segmentation. In: International conference on medical imaging with deep learning. pp. 285--296. PMLR (2019)

\bibitem{kumar2017dataset}
Kumar, N., Verma, R., Sharma, S., Bhargava, S., Vahadane, A., Sethi, A.: A dataset and a technique for generalized nuclear segmentation for computational pathology. IEEE transactions on medical imaging  \textbf{36}(7),  1550--1560 (2017)

\bibitem{landman2015miccai}
Landman, B., Xu, Z., Igelsias, J., Styner, M., Langerak, T., Klein, A.: Miccai multi-atlas labeling beyond the cranial vault--workshop and challenge. In: Proc. MICCAI Multi-Atlas Labeling Beyond Cranial Vault—Workshop Challenge. vol.~5, p.~12 (2015)

\bibitem{li2021involution}
Li, D., Hu, J., Wang, C., Li, X., She, Q., Zhu, L., Zhang, T., Chen, Q.: Involution: Inverting the inherence of convolution for visual recognition. In: Proceedings of the IEEE/CVF conference on computer vision and pattern recognition. pp. 12321--12330 (2021)

\bibitem{linguraru_swin-umamba_2024}
Liu, J., Yang, H., Zhou, H.Y., Xi, Y., Yu, L., Li, C., Liang, Y., Shi, G., Yu, Y., Zhang, S., Zheng, H., Wang, S.: Swin-{UMamba}: Mamba-based {UNet} with {ImageNet}-based pretraining. In: Linguraru, M.G., Dou, Q., Feragen, A., Giannarou, S., Glocker, B., Lekadir, K., Schnabel, J.A. (eds.) Medical Image Computing and Computer Assisted Intervention – {MICCAI} 2024, vol. 15009, pp. 615--625. Springer Nature Switzerland (2024). \doi{10.1007/978-3-031-72114-4_59}, series Title: Lecture Notes in Computer Science

\bibitem{liu2024vmamba}
Liu, Y., Tian, Y., Zhao, Y., Yu, H., Xie, L., Wang, Y., Ye, Q., Jiao, J., Liu, Y.: Vmamba: Visual state space model. Advances in neural information processing systems  \textbf{37},  103031--103063 (2024)

\bibitem{liu_swin_2021}
Liu, Z., Lin, Y., Cao, Y., Hu, H., Wei, Y., Zhang, Z., Lin, S., Guo, B.: Swin transformer: {Hierarchical} vision transformer using shifted windows. In: Proceedings of the {IEEE}/{CVF} {International} {Conference} on {Computer} {Vision}. pp. 10012--10022 (2021)

\bibitem{liu2021swin}
Liu, Z., Lin, Y., Cao, Y., Hu, H., Wei, Y., Zhang, Z., Lin, S., Guo, B.: Swin transformer: Hierarchical vision transformer using shifted windows. In: Proceedings of the IEEE/CVF international conference on computer vision. pp. 10012--10022 (2021)

\bibitem{U-Mamba}
Ma, J., Li, F., Wang, B.: U-mamba: Enhancing long-range dependency for biomedical image segmentation. arXiv preprint arXiv:2401.04722  (2024)

\bibitem{ma2025u}
Ma, X., Du, Y., Sui, D.: A u-shaped architecture based on hybrid cnn and mamba for medical image segmentation. Applied Sciences  \textbf{15}(14), ~7821 (2025)

\bibitem{oktay2018attention}
Oktay, O., Schlemper, J., Folgoc, L.L., Lee, M., Heinrich, M., Misawa, K., Mori, K., McDonagh, S., Hammerla, N.Y., Kainz, B., et~al.: Attention u-net: Learning where to look for the pancreas. arXiv preprint arXiv:1804.03999  (2018)

\bibitem{radosavovic2020designing}
Radosavovic, I., Kosaraju, R.P., Girshick, R., He, K., Doll{\'a}r, P.: Designing network design spaces. In: Proceedings of the IEEE/CVF conference on computer vision and pattern recognition. pp. 10428--10436 (2020)

\bibitem{rahman2024multi}
Rahman, M.M., Marculescu, R.: Multi-scale hierarchical vision transformer with cascaded attention decoding for medical image segmentation. In: Medical Imaging with Deep Learning. pp. 1526--1544. PMLR (2024)

\bibitem{rahman2024emcad}
Rahman, M.M., Munir, M., Marculescu, R.: Emcad: Efficient multi-scale convolutional attention decoding for medical image segmentation. In: Proceedings of the IEEE/CVF Conference on Computer Vision and Pattern Recognition. pp. 11769--11779 (2024)

\bibitem{ronneberger2015u}
Ronneberger, O., Fischer, P., Brox, T.: U-net: Convolutional networks for biomedical image segmentation. In: International Conference on Medical image computing and computer-assisted intervention. pp. 234--241. Springer (2015)

\bibitem{ruan2024vm}
Ruan, J., Li, J., Xiang, S.: Vm-unet: Vision mamba unet for medical image segmentation. ACM Transactions on Multimedia Computing, Communications and Applications  (2024)

\bibitem{sirinukunwattana2017gland}
Sirinukunwattana, K., Pluim, J.P., Chen, H., Qi, X., Heng, P.A., Guo, Y.B., Wang, L.Y., Matuszewski, B.J., Bruni, E., Sanchez, U., et~al.: Gland segmentation in colon histology images: The glas challenge contest. Medical image analysis  \textbf{35},  489--502 (2017)

\bibitem{takikawa2019gated}
Takikawa, T., Acuna, D., Jampani, V., Fidler, S.: Gated-scnn: Gated shape cnns for semantic segmentation. In: Proceedings of the IEEE/CVF international conference on computer vision. pp. 5229--5238 (2019)

\bibitem{tan2019efficientnet}
Tan, M., Le, Q.: Efficientnet: Rethinking model scaling for convolutional neural networks. In: International conference on machine learning. pp. 6105--6114. PMLR (2019)

\bibitem{wang2022uctransnet}
Wang, H., Cao, P., Wang, J., Zaiane, O.R.: Uctransnet: rethinking the skip connections in u-net from a channel-wise perspective with transformer. In: Proceedings of the AAAI conference on artificial intelligence. vol.~36, pp. 2441--2449 (2022)

\bibitem{UCTransNet}
Wang, H., Cao, P., Wang, J., Zaiane, O.R.: Uctransnet: Rethinking the skip connections in u-net from a channel-wise perspective with transformer. Proceedings of the AAAI Conference on Artificial Intelligence  \textbf{36}(3),  2441--2449 (Jun 2022). \doi{10.1609/aaai.v36i3.20144}, \url{https://ojs.aaai.org/index.php/AAAI/article/view/20144}

\bibitem{wang2022mixed}
Wang, H., Xie, S., Lin, L., Iwamoto, Y., Han, X.H., Chen, Y.W., Tong, R.: Mixed transformer u-net for medical image segmentation. In: ICASSP 2022-2022 IEEE International Conference on Acoustics, Speech and Signal Processing (ICASSP). pp. 2390--2394. IEEE (2022)

\bibitem{wang2022pvt}
Wang, W., Xie, E., Li, X., Fan, D.P., Song, K., Liang, D., Lu, T., Luo, P., Shao, L.: Pvt v2: Improved baselines with pyramid vision transformer. Computational visual media  \textbf{8}(3),  415--424 (2022)

\bibitem{wang2024mamba}
Wang, Z., Zheng, J.Q., Zhang, Y., Cui, G., Li, L.: Mamba-unet: Unet-like pure visual mamba for medical image segmentation. arXiv preprint arXiv:2402.05079  (2024)

\bibitem{woo2018cbam}
Woo, S., Park, J., Lee, J.Y., Kweon, I.S.: Cbam: Convolutional block attention module. In: Proceedings of the European conference on computer vision (ECCV). pp. 3--19 (2018)

\bibitem{xu2024hc}
Xu, J.: Hc-mamba: Vision mamba with hybrid convolutional techniques for medical image segmentation. arXiv preprint arXiv:2405.05007  (2024)

\bibitem{yao2024cnn}
Yao, W., Bai, J., Liao, W., Chen, Y., Liu, M., Xie, Y.: From cnn to transformer: A review of medical image segmentation models. Journal of Imaging Informatics in Medicine  \textbf{37}(4),  1529--1547 (2024)

\bibitem{yu2019robust}
Yu, S., Xiao, D., Frost, S., Kanagasingam, Y.: Robust optic disc and cup segmentation with deep learning for glaucoma detection. Computerized Medical Imaging and Graphics  \textbf{74},  61--71 (2019)

\bibitem{zhang2025advances}
Zhang, J., Chen, X., Yang, B., Guan, Q., Chen, Q., Chen, J., Wu, Q., Xie, Y., Xia, Y.: Advances in attention mechanisms for medical image segmentation. Computer Science Review  \textbf{56},  100721 (2025)

\bibitem{zhang2023st}
Zhang, J., Qin, Q., Ye, Q., Ruan, T.: St-unet: Swin transformer boosted u-net with cross-layer feature enhancement for medical image segmentation. Computers in Biology and Medicine  \textbf{153},  106516 (2023)

\bibitem{zhou_unet_2018}
Zhou, Z., Rahman~Siddiquee, M.M., Tajbakhsh, N., Liang, J.: {UNet}++: {A} {Nested} {U}-{Net} {Architecture} for {Medical} {Image} {Segmentation}. In: Stoyanov, D., Taylor, Z., Carneiro, G.e.a. (eds.) Deep {Learning} in {Medical} {Image} {Analysis} and {Multimodal} {Learning} for {Clinical} {Decision} {Support}. pp. 3--11. Springer International Publishing, Cham (2018)

\end{thebibliography}

\end{document}